\documentclass[final]{cvpr}

\usepackage{times}
\usepackage{epsfig}
\usepackage{graphicx}
\usepackage{amsmath}
\usepackage{amssymb}

\usepackage{algorithm}
\usepackage{algorithmic}
\usepackage{multirow}
\usepackage{verbatim}
\usepackage{color}
\usepackage{gensymb}
\usepackage{amsmath}
\usepackage{subfiles}
\usepackage{bbm}
\usepackage{amssymb}  

\usepackage[switch]{lineno}

\usepackage[pagebackref=true,breaklinks=true,colorlinks,bookmarks=false]{hyperref}

\def\cvprPaperID{801} 
\def\confYear{CVPR 2021}
\begin{document}

\title{SE-SSD: Self-Ensembling Single-Stage Object Detector From Point Cloud}

\author{Wu Zheng\quad Weiliang Tang\quad Li Jiang\quad Chi-Wing Fu\\
The Chinese University of Hong Kong\\
{\tt\small \{wuzheng, lijiang, cwfu\}@cse.cuhk.edu.hk\quad  tangwl123@foxmail.com}}

\maketitle
\newcommand{\TODO}[1]{{\color{red}{[TODO: #1]}}}
\newcommand{\phil}[1]{{\color[rgb]{0.3,0.7,0.3}{[PH: #1]}}}
\newcommand{\zw}[1]{{\color[rgb]{0.7,0.3,0.7}{[ZW: #1]}}}
\newcommand{\para}[1]{\vspace{.05in}\noindent\textbf{#1}}
\def\ie{\emph{i.e.}}
\def\eg{\emph{e.g.}}
\def\etal{{\em et al.}}
\def\etc{{\em etc.}}

\ifx\allfiles\undefined
\documentclass[letterpaper]{article}
\begin{document}
\else
\chapter{abstraction}
\fi

\begin{abstract}
We present Self-Ensembling Single-Stage object Detector (SE-SSD) for accurate and efficient 3D object detection in outdoor point clouds.
Our key focus is on exploiting both soft and hard targets with our formulated constraints to jointly optimize the model, without introducing extra computation in the inference.
Specifically, SE-SSD contains a pair of teacher and student SSDs, in which we design an effective IoU-based matching strategy to filter soft targets from the teacher and formulate a consistency loss to align student predictions with them.
Also, to maximize the distilled knowledge for ensembling the teacher, we design a new augmentation scheme to produce shape-aware augmented samples to train the student, aiming to encourage it to infer complete object shapes.
Lastly, to better exploit hard targets, we design an ODIoU loss to supervise the student with constraints on the predicted box centers and orientations.
Our SE-SSD attains top performance compared with all prior published works.
Also, it attains top precisions for car detection in the KITTI benchmark (ranked 1$^{st}$ and 2$^{nd}$ on the BEV and 3D leaderboards\footnote{On the date of CVPR deadline,~\ie, Nov 16, 2020}, respectively) with an ultra-high inference speed.
The code is available at \url{https://github.com/Vegeta2020/SE-SSD}.
\end{abstract}

\ifx\allfiles\undefined
\end{document}
\fi 
\ifx\allfiles\undefined
\documentclass[letterpaper]{article}
\begin{document}
\else
\chapter{introduction}
\fi

\vspace*{-0.5mm}
\section{Introduction}

To support autonomous driving, 3D point clouds from LiDAR sensors are often adopted to detect objects near the vehicle.
This is a robust approach, since point clouds are readily available regardless of the weather (fog vs. sunny) and time of the day (day vs. night).
Hence, various point-cloud-based 3D detectors have been proposed recently.

\begin{figure}
\centering
\includegraphics[width=0.97\columnwidth]{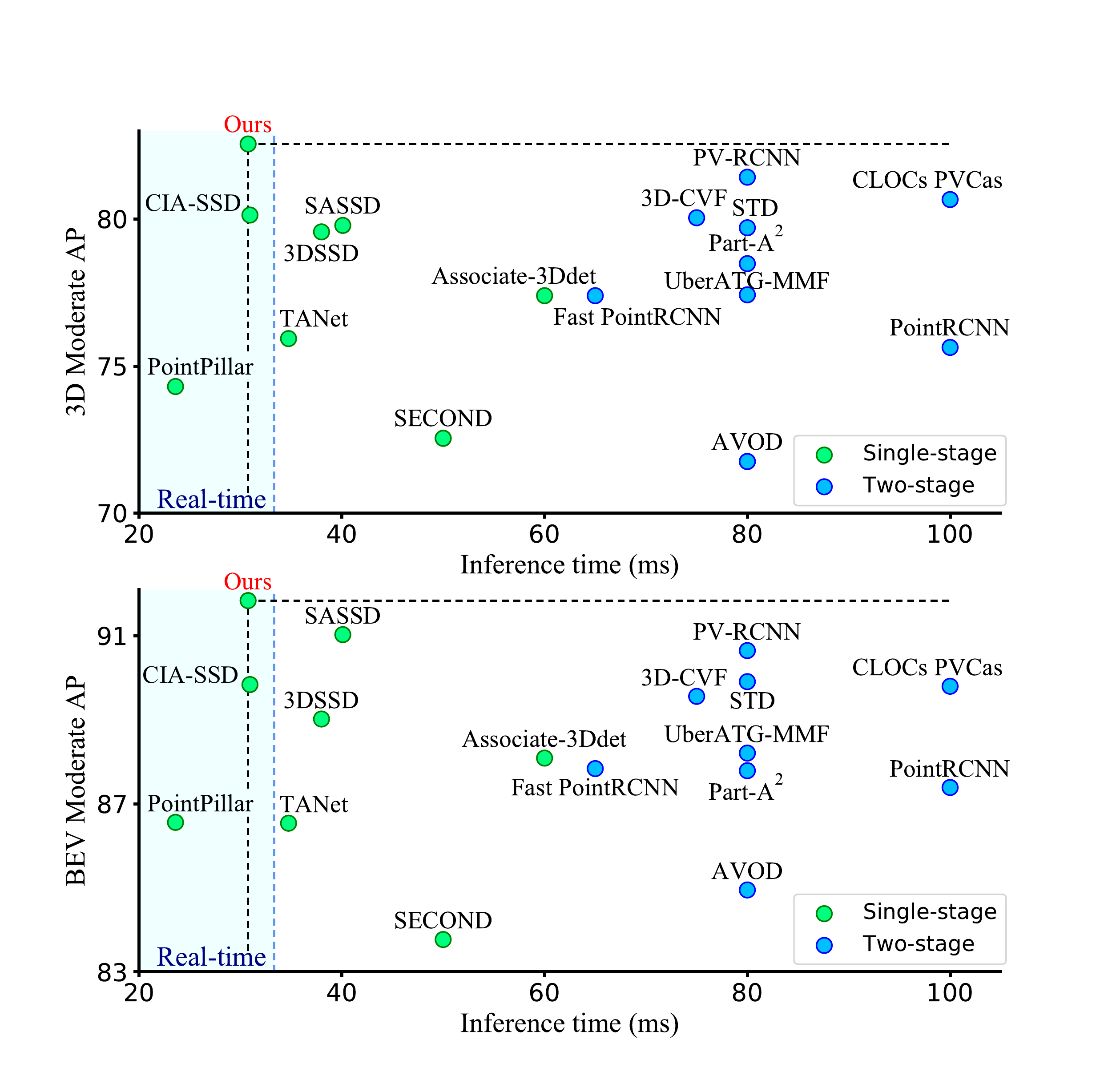}
\caption{Our SE-SSD attains top precisions on both 3D and BEV car detection in KITTI benchmark~\protect\cite{geiger2013vision} with real-time speed ($30.56$ ms), clearly outperforming all state-of-the-art detectors. Please refer to Table~\ref{table1} for a detailed comparison with more methods.}
\label{fig:cover}
\vspace*{-2.5mm}
\end{figure}

To boost the detection precision, an important factor is the quality of the extracted features.
This applies to both single-stage and two-stage detectors.
For example, the series of works~\cite{shi2019pointrcnn,Chen2019fastpointrcnn,shi2020points,shi2020pv} focus on improving the region-proposal-aligned features for a better refinement with a second-stage network.
Also, many methods~\cite{MV3D,AVOD,xie2020pi,liang2019multi,yoo20203d,pang2020clocs} try to extract more discriminative multi-modality features by fusing RGB images and 3D point clouds.
For single-stage detectors, Point-GNN~\cite{shi2020point} adapts a graph neural network to obtain a more compact representation of point cloud, while TANet~\cite{liu2020tanet} designs a delicate triple attention module to consider the feature-wise relation.
Though these approaches give significant insights, the delicate designs are often complex and could slow down the inference, especially for the two-stage detectors.

To meet the practical need, especially in autonomous driving, 3D object detection demands high efficiency on top of high precision.
Hence, another stream of works,~\eg, SA-SSD~\cite{he2020structure} and Associate-3Ddet~\cite{du2020associate}, aim to exploit auxiliary tasks or further constraints to improve the feature representation, without introducing additional computational overhead during the inference.
Following this stream of works, we formulate the Self-Ensembling Single-Stage object Detector (SE-SSD) to address the challenging 3D detection task based only on LiDAR point clouds.

To boost the detection precision, while striving for high efficiency, we design the SE-SSD framework with a pair of teacher SSD and student SSD, as inspired by~\cite{tarvainen2017mean}.
The teacher SSD is ensembled from the student.
It produces relatively more precise bounding boxes and confidence, which serve as soft targets to supervise the student.
Compared with manually-annotated hard targets (labels), soft targets from the teacher often have higher entropy, thus offering more information~\cite{hinton2015distilling} for the student to learn from.
Hence, we exploit both soft and hard targets with our formulated constraints to jointly optimize the model, while incurring no extra inference time.
To encourage the bounding boxes and confidence predicted by the student to better align with the soft targets, we design an effective IoU-based matching strategy to filter soft targets and pair them with student predictions, and further formulate a consistency loss to reduce the misalignment between them.

On the other hand, to enable the student SSD to effectively explore a larger data space, we design a new augmentation scheme on top of conventional augmentation strategies to produce augmented object samples in a shape-aware manner.
By this scheme, we can encourage the model to infer the complete object shape from incomplete information.
It is also a plug-and-play and general module for 3D detectors.
Furthermore, hard targets are still essential in the supervised training, as they are the final targets for the model convergence.
To better exploit them, we formulate a new orientation-aware distance-IoU (ODIoU) loss to supervise the student with constraints on both the center and orientation of the predicted bounding boxes.
Overall, our SE-SSD is trained in a fully-supervised manner to best boost the detection performance, in which all the designed modules are needed only in the training, so there are no extra computation during the inference.

In summary, our contributions include
(i) the Self-Ensembling Single-Stage object Detector (SE-SSD) framework, optimized by our formulated consistency constraint to better align predictions with the soft targets;
(ii) a new augmentation scheme to produce shape-aware augmented ground-truth objects; and
(iii) an Orientation-aware Distance-IoU (ODIoU) loss to supervise the detector using hard targets.
Our SE-SSD attains state-of-the-art performance on both 3D and BEV car detection in the KITTI benchmark~\cite{geiger2013vision} and demonstrates ultra-high inference speed (32 FPS) on commodity CPU-GPU, clearly outperforming all prior published works, as presented in Figure~\ref{fig:cover}.

\ifx\allfiles\undefined
\end{document}
\fi

\ifx\allfiles\undefined
\documentclass[letterpaper]{article}
\begin{document}
\else
\chapter{related_work}
\fi

\section{Related Work}

In general, 3D detectors are categorized into two types:
(i) {\em single-stage detectors\/} regress bounding box and confidence directly from features learned from the input, and
(ii) {\em two-stage detectors\/} use a second stage to refine the first-stage predictions with region-proposal-aligned features.
So, two-stage detectors often attain higher precisions benefited from the extra stage,
whereas single-stage detectors usually run faster due to simpler network structures.
Recent trend (see Figure~\ref{fig:cover} and Table~\ref{table1}) reveals that the precisions of single-stage detectors~\cite{he2020structure,yang20203dssd} gradually approach those of the two-stage detectors~\cite{shi2020pv,shi2020points,yang2019std}.
This motivates us to focus this work on developing a single-stage detector and {\em aim for both high precision and high speed\/}.

\textbf{Two-stage Object Detectors}
Among these two-stage detectors, PointRCNN~\cite{shi2019pointrcnn} uses PointNet~\cite{qi2017pointnet} to fuse semantic features and raw points from region proposals for a second-stage refinement.
Part-$A^2$~\cite{shi2020points} exploits a voxel-based network to extract region proposal features to avoid ambiguity and further improve the feature representation.
Similarly, STD~\cite{yang2019std} converts region-proposal semantic features to a compact representation with voxelization and reduce the number of anchors to improve the performance.
PV-RCNN~\cite{shi2020pv} utilizes both point-based and voxel-based networks to extract features from the voxels and raw points inside the region proposal.
3D-CVF~\cite{yoo20203d} obtains semantics from multi-view images and fuses them with point features in both stages,
whereas CLOCs PVCas~\cite{pang2020clocs} fuses semantic features from images and points to refine the predicted confidence.

\begin{figure*}
\centering
\includegraphics[width=16.25cm]{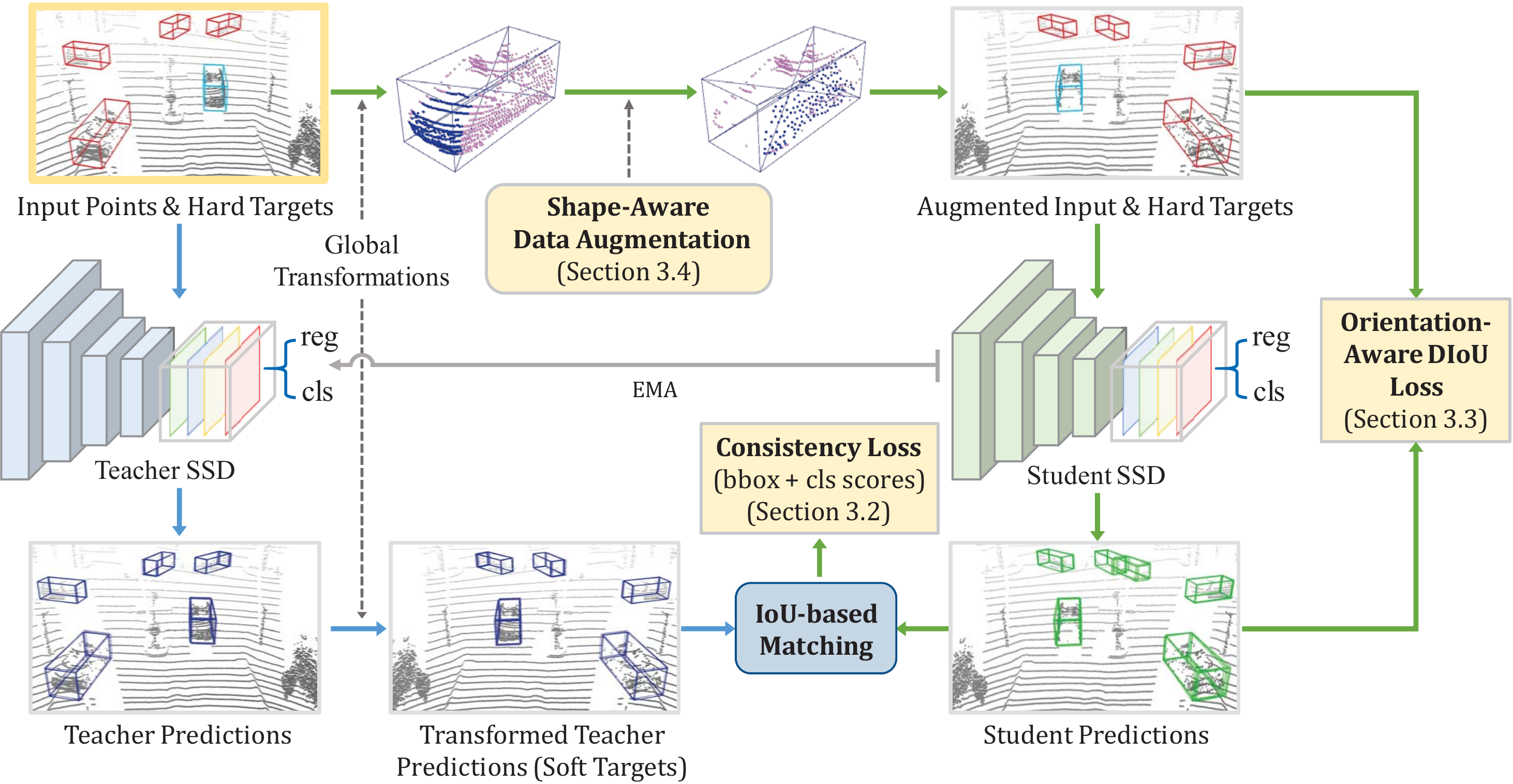}
\vspace*{-1mm}
\caption{
The framework of our Self-Ensembling Single-Stage object Detector (SE-SSD) with a teacher SSD and a student SSD.
To start, we feed the input point cloud to the teacher to produce relatively precise bounding boxes and confidence, and take these predictions (after global transformations) as soft targets to supervise the student with our {\em consistency loss\/} (Section~\ref{sec:3.3}).
On the top branch, we apply the same global transformations to the input, then perform our {\em shape-aware data augmentation\/} (Section~\ref{sec:3.2}) to generate augmented samples as inputs to the student.
Further, we formulate the {\em Orientation-aware Distance-IoU loss\/} (Section~\ref{sec:3.4}) to supervise the student with hard targets, and update the teacher parameters based on the student parameters with the exponential moving average (EMA) strategy.
In this way, the framework can boost the precisions of the detector significantly without incurring extra computation during the inference.
}
\label{pipeline}
\vspace*{-2.5mm}
\end{figure*}

\textbf{Single-stage Object Detectors}
VoxelNet~\cite{zhou2018voxelnet} proposes the voxel feature encoding layer to extract features from point clouds.
PointPillar~\cite{lang2019pointpillars} divides a point cloud into pillars for efficient feature learning.
SECOND~\cite{yan2018second} modifies the sparse convolution~\cite{graham20183d,liu2015sparse} to efficiently extract features from sparse voxels.
TANet~\cite{liu2020tanet} proposes the triple attention module to consider feature-wise relation in the feature extraction.
Point-GNN~\cite{shi2020point} exploits a graph neural network to learn point features.
3DSSD~\cite{yang20203dssd} combines feature- and point-based sampling to improve the classification.
Associate-3Ddet~\cite{du2020associate} extracts feature from complete point clouds to supervise the features learned from incomplete point clouds, encouraging the model to infer from incomplete points.
SA-SSD~\cite{he2020structure} adopts an auxiliary network in parallel with the backbone to regress box centers and semantic classes to enrich the features.
CIA-SSD~\cite{he2020structure} adopts a lightweight BEV network for extraction of robust spatial-semantic features, combined with an IoU-aware confidence rectification and DI-NMS for better post processing.
Inspired by ~\cite{tarvainen2017mean}, SESS~\cite{zhao2020sess} addresses the detection in indoor scenes with a semi-supervised strategy to reduce the dependence on manual annotations.

Different from prior works, we aim to exploit both soft and hard targets to refine features in a fully-supervised manner through our novel constraints and augmentation scheme.
Also, compared with {\em all prior\/} single- and two-stage detectors, our SE-SSD attains the {\em highest average precisions for both 3D and BEV car detection\/} in the KITTI benchmark~\cite{geiger2013vision} and exhibits {\em very high efficiency\/}.

\ifx\allfiles\undefined
\end{document}
\fi 
\ifx\allfiles\undefined
\documentclass[letterpaper]{article}
\begin{document}
\else
\chapter{framework}
\fi

\section{Self-Ensembling Single Stage Detector}

\subsection{Overall Framework}
\label{sec:3.1}

Figure~\ref{pipeline} shows the framework of our self-ensembling single-stage object detector (SE-SSD), which has a teacher SSD (left) and a student SSD (right).
Different from prior works on outdoor 3D object detection, we simultaneously employ and train two SSDs (of same architecture), such that the student can explore a larger data space via the augmented samples and be better optimized with the associated soft targets predicted by the teacher.
To train the whole SE-SSD, we first initialize both the teacher and student with a pre-trained SSD model.
Then, started from an input point cloud, our framework has two processing paths:
\begin{itemize}
\vspace*{-1mm}
\item
In the first path (blue arrows in Figure~\ref{pipeline}), the teacher produces relatively precise predictions from the raw input point cloud.
Then, we apply a set of global transformations on the prediction results and take them as soft targets to supervise the student SSD.
\vspace*{-1mm}
\item
In the second path (green arrows in Figure~\ref{pipeline}), we perturb the same input by the same global transformations as in the first path plus our shape-aware data augmentation (Section~\ref{sec:3.2}).
Then, we feed the augmented input to the student, and
train it with
(i) our consistency loss (Section~\ref{sec:3.3}) to align the student predictions with the soft targets; and
(ii) when we augment the input, we bring along its hard targets (Figure~\ref{pipeline} (top right)) to supervise the student with our orientation-aware distance-IoU loss (Section~\ref{sec:3.4}).
\end{itemize}

\vspace*{-1mm}
In the training, we iteratively update the two SSD models: optimize the student with the above two losses and update the teacher using only the student parameters by a standard exponential moving average (EMA).
Thus, the teacher can obtain distilled knowledge from student and produce soft targets to supervise student.
So, we call the final trained student a {\em self-ensembling single-stage object detector\/}.

\vspace*{-3.5mm}
\paragraph{Architecture of Teacher \& Student SSD}
The model has the same structure as~\cite{zheng2020cia} to efficiently encode point clouds, but we remove the Confidence
Function and DI-NMS.
It has a sparse convolution network (SPConvNet), a BEV convolution network (BEVConvNet), and a multi-task head (MTHead).
BEV means bird's eye view.
After point cloud voxelization, we find mean 3D coordinates and point density per voxel as the initial feature, then extract features using SPConvNet, which has four blocks (\{2, 2, 3, 3\} submanifold sparse convolution~\cite{graham20183d} layers) with a sparse convolution~\cite{liu2015sparse} layer at the end.
Next, we concatenate the sparse 3D feature along $z$ into 2D dense ones for feature extraction with the BEVConvNet.
Lastly, we use MTHead to regress bounding boxes and perform classification.


\subsection{Consistency Loss}
\label{sec:3.3}
In 3D object detection, the patterns of point clouds in pre-defined anchors may vary significantly due to distances and different forms of object occlusion.
Hence, samples of the {\em same\/} hard target may have very different point patterns and features.
In contrast, soft targets can be more informative per training sample, and helps to reveal the difference between data samples of the same class~\cite{hinton2015distilling}.
This motivates us to treat the relatively more precise teacher predictions as soft targets and employ them to jointly optimize the student with hard targets.
Accordingly, we formulate a consistency loss to optimize the student network with soft targets.

We first design an effective IoU-based matching strategy before calculating the consistency loss, aiming to pair the
non-axis-aligned teacher and student boxes predicted from the very sparse outdoor point clouds.
To obtain high-quality soft targets from the teacher, we first filter out those predicted bounding boxes (for both teacher and student) with confidence less than threshold $\tau_{c}$, which helps reduce the calculation of the consistency loss.
Next, we calculate the IoU between every pair of remaining student and teacher bounding boxes, and filter out the pairs with IoUs less than threshold ${\tau_{I}}$, thus avoiding to mislead the student with unrelated soft targets; We denote $N$ and $N'$ as the initial and final number of box pairs, respectively.
Thus, we keep only the highly-overlapping student-teacher pairs.
Lastly, for each student box, we pair it with the teacher bounding box that has the largest IoU with it, aiming to increase the confidence of the soft targets.
Compared with hard targets, the filtered soft targets are often closer to the student predictions, as they are predicted based on similar features.
So, soft targets can better guide the student to fine-tune the predictions and reduce the gradient variance for better training.

Different from the IoU loss, Smooth-$L_1$ loss~\cite{liu2016ssd} can evenly treat all dimensions in the predictions, without biasing toward any specific one, so the features corresponding to different dimensions can also be evenly optimized.
Hence, we adopt it to formulate our consistency loss for bounding boxes ($\mathcal L^{c}_{box}$) to minimize the misalignment errors between each pair of teacher and student bounding boxes:
\begin{equation}\label{loc_cons}
    \begin{split}
      &\mathcal L^{c}_{box} = \frac{1}{N'}\sum_{i=1}^{N}\mathbbm{1}(IoU_{i} > \tau_{I})\sum_{e} \frac{1}{7} \mathcal{L}^{c}_{\delta_e} \\
        \text{and} \ \
        &\delta_{e}=
          \begin{cases}
              |e_{s}-e_{t}|  & \textnormal{if} \  e\in \{x, y, z, w, l, h\} \\
              |sin(e_{s}-e_{t})| & \textnormal{if} \ e\in \{r\} \
          \end{cases}
    \end{split}
\end{equation}
where
$\{x, y, z\}$, $\{w, l, h\}$, and ${r}$ denote the center position, sizes, and orientation of a bounding box, respectively, predicted by teacher (subscript $t$) or student (subscript $s$),
$\mathcal{L}^{c}_{\delta_e}$ denotes the Smooth-$L_1$ loss of $\delta_e$,
and
$IoU_i$ denotes the largest IoU of the $i$-th student bounding box with all teacher bounding boxes.
Next, we formulate the consistency loss for classification score ($\mathcal L^{c}_{cls}$) to minimize the difference in predicted confidence of student and teacher:
\vspace*{-2mm}
\begin{equation}\label{cls_cons}
    \begin{split}
        &\mathcal L^{c}_{cls} = \frac{1}{N'}\sum_{i=1}^{N}\mathbbm{1}(IoU_{i} > \tau_{I}) \mathcal{L}^{c}_{\delta_{c}} \\
        \text{and} \ \
        &\delta_{c}=|\sigma(c_s) - \sigma(c_t)| \
    \end{split}
\end{equation}
%
where $\mathcal{L}^{c}_{\delta_c}$ denotes the Smooth-$L_1$ loss of $\delta_c$, and $\sigma(c_s)$ and $\sigma(c_t)$ denote the sigmoid classification scores of student and teacher, respectively.
Here, we adopt the sigmoid function to normalize the two predicted confidences, such that the deviation between the normalized values can be kept inside a small range.
Combining Eqs~\eqref{loc_cons} and~\eqref{cls_cons}, we can obtain the overall consistency loss as
\begin{equation}\label{over_cons}
        \mathcal L_{cons} = \mathcal L^{c}_{cls} + \mathcal L^{c}_{box} \
\end{equation}
where we empirically set the same weight for both terms.

\begin{figure}[!t]
\centering
\includegraphics[width=6.2cm]{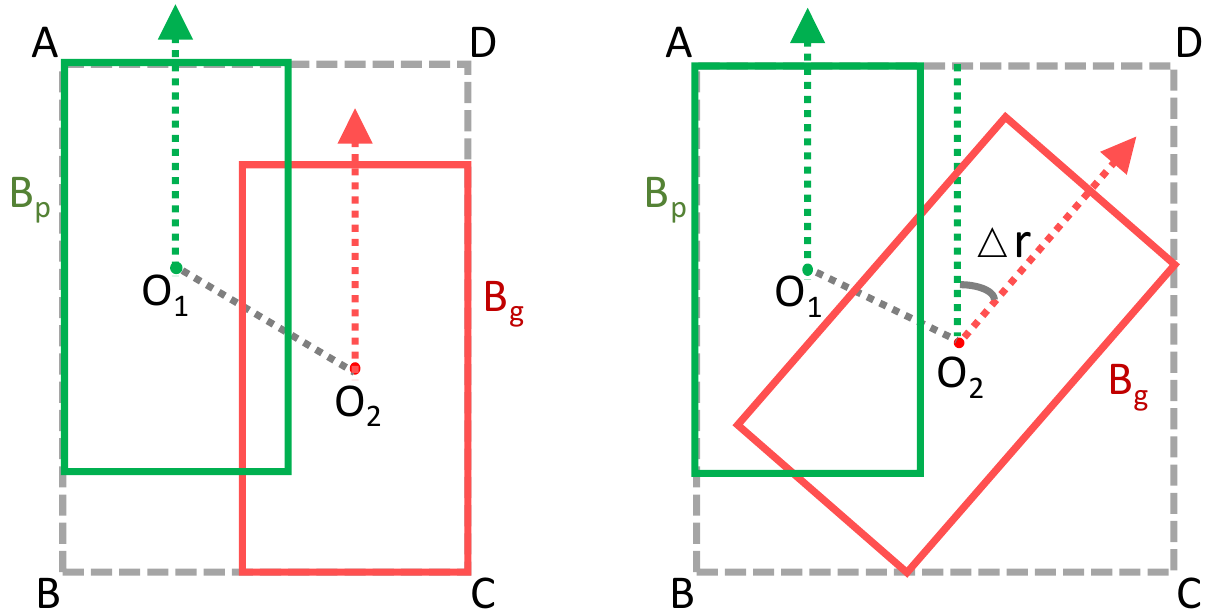}
\caption{Illustrating axis-aligned bounding boxes (left) and non-axis-aligned bounding boxes (right) in Bird's Eye View (BEV), where the red and green boxes represent the ground truth and predicted box, respectively.
To handle non-axis-aligned bounding boxes (right), our ODIoU loss impose constraints on both the box center distance
$|O_{1}O_{2}|$ and the orientation difference $\triangle r$ in BEV.}
\label{fig:IoUloss}
\end{figure}

\subsection{Orientation-Aware Distance-IoU Loss}
\label{sec:3.4}

In supervised training with hard targets, Smooth-$L_1$ loss~\cite{liu2016ssd} is often adopted to constrain the bounding box regression.
However, due to long distances and occlusion in outdoor scenes, it is hard to acquire sufficient information from the sparse points to precisely predict all dimensions of the bounding boxes.
To better exploit hard targets for regressing bounding boxes, we design the Orientation-aware Distance-IoU loss (ODIoU) to
focus more attention on the alignment of box centers and orientations between the predicted and ground-truth bounding boxes; see Figure~\ref{fig:IoUloss}.

Inspired by~\cite{zheng2020distance}, we impose a constraint on the distance between the 3D centers of the predicted and ground-truth bounding boxes to minimize the center misalignment.
More importantly, we design a novel orientation constraint on the predicted BEV angle, aiming to further minimize the orientation difference between the predicted and ground-truth boxes.
In 3D object detection, such a constraint is significant for the precise alignment between the non-axis-aligned boxes in the bird's eye view (BEV).
Also, we empirically find that this constraint is an important means to further boost the detection precision.
Compared with Smooth-$L_1$ loss, our ODIoU loss enhances the alignment of box centers and orientations, which are easy to infer from the points distributed on the object surface, thus leading to a better performance.
Overall, our ODIoU loss is formulated as
\vspace{-1mm}
\begin{equation}\label{diou}
        \mathcal L^{s}_{box} = 1 - IoU(B_p, B_g) + \frac{c^2}{d^2} + \gamma(1-|cos(\triangle r)|) \
\end{equation}
where
$B_p$ and $B_g$ denote the predicted and ground-truth bounding boxes, respectively,
$c$ denotes the distance between the 3D centers of the two bounding boxes (see $|O_1O_2|$ in Figure~\ref{fig:IoUloss}),
$d$ denotes the diagonal length $|AC|$ of the minimum cuboid that encloses both bounding boxes;
$\triangle r$ denotes the BEV orientation difference between $B_p$ and $B_g$; and
$\gamma$ is a hyper-parameter weight.

\begin{figure}[!t]
\centering
\includegraphics[width=7.5cm]{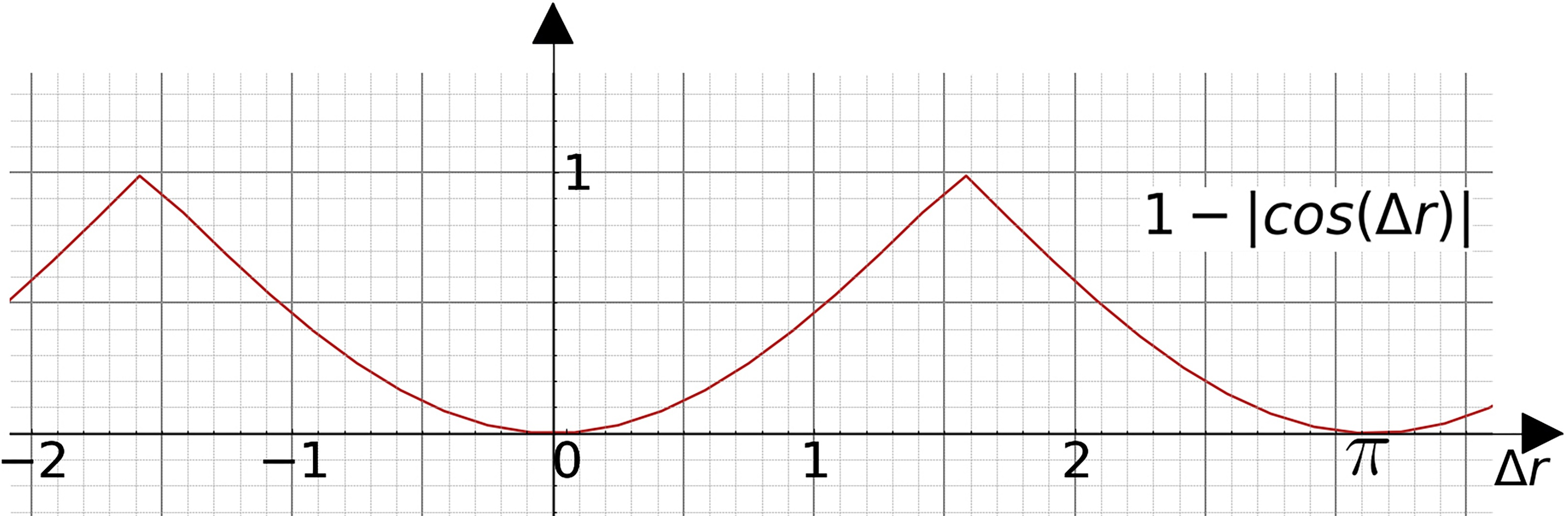}
\caption{We formulate the orientation constraint $(1-|cos(\triangle r)|)$ in the ODIoU loss to precisely minimize the orientation difference between bounding boxes; note the gradient of the term.}
\label{fig:plot_deltaR}
\vspace{-2mm}
\end{figure}

In our ODIoU loss formula, $(1-|cos(\triangle r)|)$ is an important term we designed specifically to encourage the predicted bounding box to rotate to the nearest direction that is parallel to the ground-truth orientation.
When $\triangle r$ equals $\frac{\pi}{2}$ or $-\frac{\pi}{2}$,~\ie, the orientations of the two boxes are perpendicular to each other, so the term attains its maxima.
When $\triangle r$ equals $0$, $\pi$, or $-\pi$, the term attains its minima, which is zero.
As shown in Figure~\ref{fig:plot_deltaR}, we can further look at the gradient of $(1-|cos(\triangle r)|)$.
When the training process minimizes the term, its gradient will help to bring $\triangle r$ to $0$, $\pi$, or $-\pi$, which is the nearest location to minimize the loss.
It is because the gradient magnitude is positively correlated to the angle difference, thus promoting fast convergence and smooth fine-tuning in different training stages.

Besides, we use the Focal loss~\cite{lin2017focal} and cross-entropy loss for the bounding box classification ($\mathcal{L}^{s}_{cls}$) and direction classification ($\mathcal{L}^{s}_{dir}$), respectively.
Hence, the overall loss to train the student SSD is
\vspace{-1mm}
\begin{equation}\label{totalloss}
  \mathcal{L}_\text{student}=\mathcal{L}^{s}_\text{cls}+\omega_1\mathcal{L}^s_\text{box}+\omega_2\mathcal{L}^s_\text{dir}+ \mu_t(\mathcal L^c_\text{cls} + \mathcal L^c_\text{box}) \
\end{equation}
where $\mathcal{L}^{s}_\text{box}$ is the ODIoU loss for regressing the boxes, and the loss weights $\omega_1$, $\omega_2$, and $\mu_t$ are hyperparameters.
Futher, our SSD can be pre-trained with the same settings as SE-SSD but without the consistency loss and teacher SSD.

\begin{figure}
\centering
\includegraphics[width=8cm]{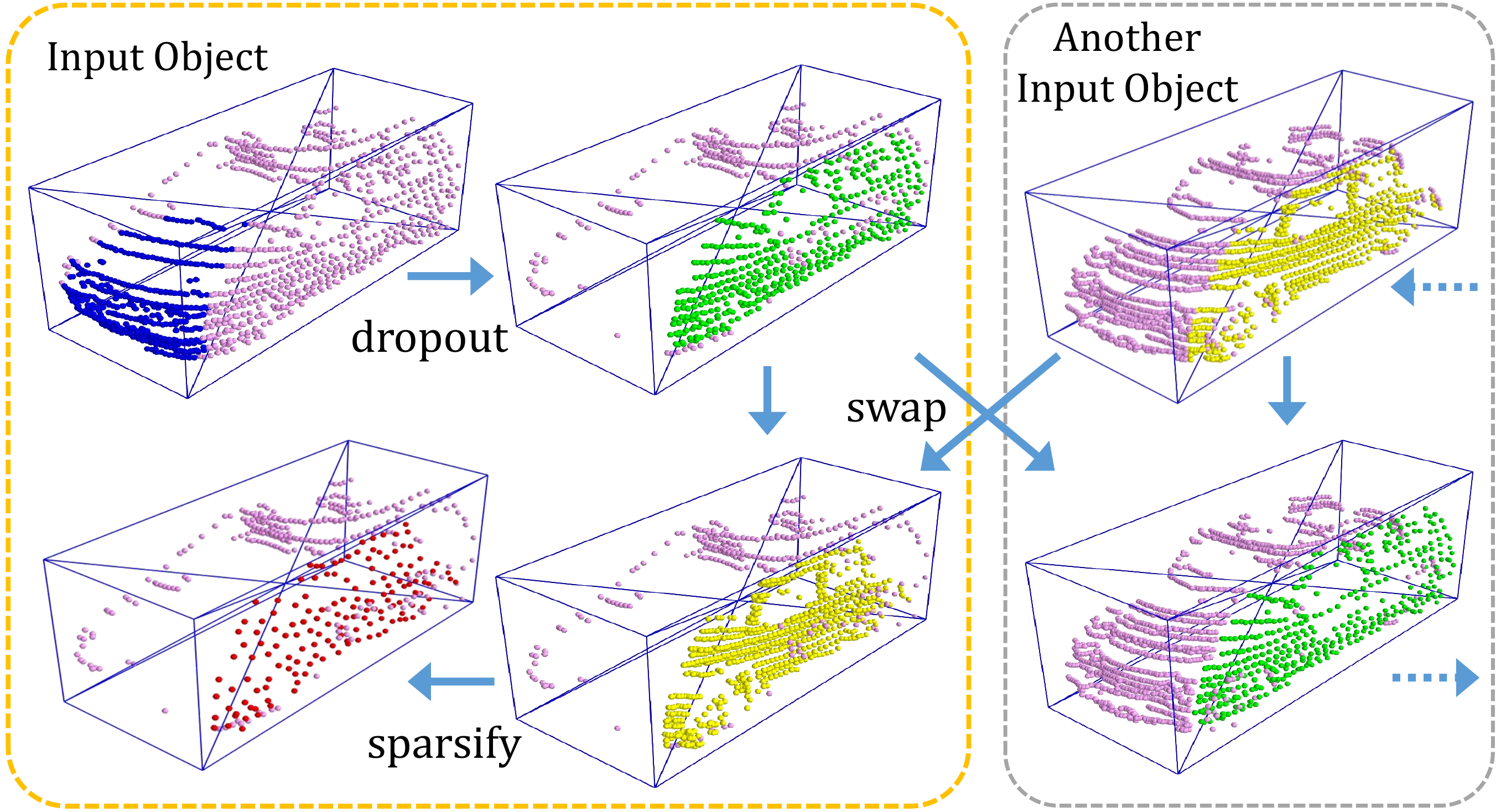}
\caption{Illustrating how the shape-aware data augmentation scheme works on a ground-truth object.
We divide the object's point cloud into six pyramidal subsets (one for each face of the object's bounding box), and then independently augment each subset by random dropout, swap, and/or sparsify operations.
}
\label{fig:sada}
\vspace*{-2mm}
\end{figure}


\subsection{Shape-Aware Data Augmentation}
\label{sec:3.2}
Data augmentation is important to improve a model's generalizability.
To enable the student SSD to explore a larger data space, we design a new shape-aware data augmentation scheme to boost the performance of our detector.
Our insight comes from the observation that the point cloud patterns of ground-truth objects could vary significantly due to occlusions, changes in distance, and diversity of object shapes in practice.
So, we design the shape-aware data augmentation scheme to mimic how point clouds are affected by these factors when augmenting the data samples.

By design, our shape-aware data augmentation scheme is a plug-and-play module. 
To start, for each object in a point cloud, we find its ground-truth bounding box centroid and connect the centroid with the box faces to form pyramidal volumes that divide the object points into six subsets.
Observing that LiDAR points are distributed mainly on object surfaces, the division is like an object disassembly, and our augmentation scheme efficiently augments each object's point cloud by manipulating these divided point subsets like disassembled parts.

In details, our scheme performs the following three operations with randomized probabilities $p_1$, $p_2$, and $p_3$, respectively:
(i) {\em random dropout\/} removes all points (blue) in a randomly-chosen pyramid (Figure~\ref{fig:sada} (top-left)), mimicking a partial object occlusion to help the network to infer a complete shape from the remained points.
(ii) {\em random swap\/} randomly selects another input object in the current scene and swap a point subset (green) with the point subset (yellow) in the same pyramid of the other input object (Figure~\ref{fig:sada} (middle)), thus increasing the diversity of object samples by exploiting the surface similarity across objects.
(iii) {\em random sparsifying\/} subsamples points in a randomly-chosen pyramid using farthest point sampling~\cite{qi2017pointnet++}, mimicking the sparsity variation of points due to changes in distance from LiDAR camera; see the sparsified points (red) in Figure~\ref{fig:sada}.

Furthermore, before the shape-aware augmentation, we perform a set of global transformations on the input point cloud, including a random translation, flipping, and scaling; see ``global transformations'' in Figure~\ref{pipeline}.

\begin{figure*}
\centering
\includegraphics[width=17.3cm]{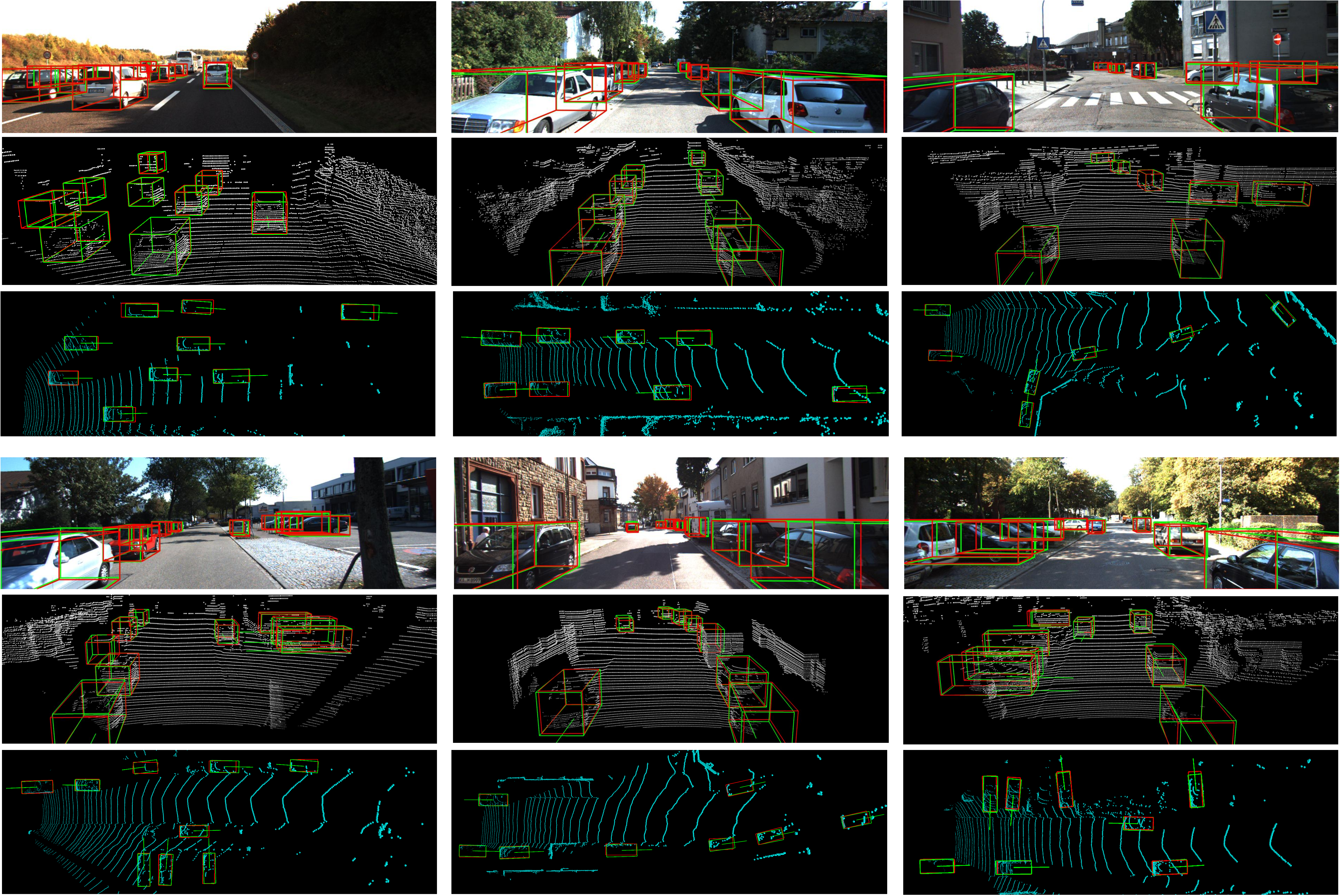}
\vspace*{2mm}
\caption{Snapshots of our 3D and BEV detection results on the KITTI validation set.
We render the predicted and ground-truth bounding boxes in green and red, respectively, and project them back onto the color images for visualization.}
\label{detresults}
\vspace*{-3mm}
\end{figure*}

\ifx\allfiles\undefined
\end{document}
\fi

\ifx\allfiles\undefined

\documentclass[letterpaper]{article}
\begin{document}
\else
\chapter{experiments}
\fi

\begin{table*}[t]
   \centering
   \footnotesize
   \begin{tabular}{c|c|c|c|cccc|cccc|c}
       \hline
       &
       \multicolumn{1}{c|}{ \multirow{2}{*}{Method}} & \multicolumn{1}{c|}{ \multirow{2}{*}{Reference}}& \multicolumn{1}{c|}{ \multirow{2}{*}{Modality}} & \multicolumn{4}{|c|}{$3D$} & \multicolumn{4}{|c|}{$BEV$} & \multicolumn{1}{|c}{\multirow{2}{*}{Time (ms)}} \\
       \cline{5-12}
       &
       \multicolumn{1}{c|}{} & \multicolumn{1}{c|}{} & \multicolumn{1}{c|}{}& \multicolumn{1}{|c}{Easy} & \multicolumn{1}{c}{Mod} & \multicolumn{1}{c}{Hard} & \multicolumn{1}{c|}{mAP} &
       \multicolumn{1}{|c}{Easy} & \multicolumn{1}{c}{Mod} & \multicolumn{1}{c}{Hard} & \multicolumn{1}{c|}{mAP} & \multicolumn{1}{|c}{} \\
       \hline
       \hline
         \parbox[t]{2mm}{\multirow{14}{*}{\rotatebox[origin=c]{90}{Two-stage}}}
         & MV3D~\cite{MV3D}     &CVPR 2017   & {LiDAR+RGB}   & 74.97  & 63.63  & 54.00 &64.2 &86.62 &78.93 &69.80 &78.45 & 360\\
         & F-PointNet~\cite{FPOINTNET} &CVPR 2018   & {LiDAR+RGB}   & 82.19  & 69.79  & 60.59 &70.86 &91.17 &84.67 &74.77 &83.54 & 170\\
         & AVOD~\cite{AVOD}            &IROS 2018   & {LiDAR+RGB}   & 83.07  & 71.76  & 65.73 &73.52 &89.75 &84.95 &78.32 &84.34 & 100\\
         & PointRCNN~\cite{shi2019pointrcnn} &CVPR 2019   & {LiDAR}     & 86.96  & 75.64  & 70.70 &77.77 &92.13 &87.39 &82.72 & 87.41 & 100\\
         & F-ConvNet~\cite{wang2019frustum}  &IROS 2019   & {LiDAR+RGB}   & 87.36  & 76.39  & 66.69 &76.81 &91.51 &85.84 &76.11 &84.49 & 470*\\
         & 3D IoU Loss~\cite{zhou2019iou}    &3DV 2019   & {LiDAR}     & 86.16  & 76.50  & 71.39 &78.02 &91.36  &86.22  &81.20  &86.26 & 80*\\
         & Fast PointRCNN~\cite{Chen2019fastpointrcnn} &ICCV 2019   & {LiDAR}     & 85.29  & 77.40  & 70.24 &77.64 &90.87 &87.84 &80.52 &86.41 & 65\\
         & UberATG-MMF~\cite{liang2019multi} &CVPR 2019   & {LiDAR+RGB}   & 88.40  & 77.43  & 70.22 &78.68 &93.67 &88.21 &81.99 &87.96 & 80\\
         & Part-$A^2$~\cite{shi2020points}   &TPAMI 2020   & {LiDAR}     & 87.81  & 78.49  & 73.51 &79.94 &91.70  &87.79  &84.61  &88.03 & 80\\
         & STD~\cite{yang2019std}            &ICCV 2019   & {LiDAR}     & 87.95  & 79.71  & 75.09 &80.92 &94.74 &89.19 &86.42 & 90.12 & 80\\
         & 3D-CVF~\cite{yoo20203d}           &ECCV 2020   & {LiDAR+RGB}   & 89.20  & 80.05  & 73.11 &80.79 &93.52  &89.56 &82.45  &88.51 & 75\\
         & CLOCs PVCas~\cite{pang2020clocs}  &IROS 2020   & {LiDAR+RGB}     & 88.94  & 80.67  &77.15 &82.25  &93.05  &89.80  &86.57  &89.81 &100*\\
         & PV-RCNN~\cite{shi2020pv}          &CVPR 2020   & {LiDAR}     & 90.25  & 81.43  & 76.82 & 82.83 &94.98  &90.65  &86.14  &90.59 & 80*\\
         & De-PV-RCNN~\cite{2020deformable} &ECCVW 2020   & {LiDAR}     & 88.25  & 81.46  & 76.96 &82.22 &92.42  &90.13  &85.93  &89.49 & 80*\\
      \hline
      \hline
         \parbox[t]{2mm}{\multirow{12}{*}{\rotatebox[origin=c]{90}{One-stage}}}
         & VoxelNet~\cite{zhou2018voxelnet} &CVPR 2018   &{LiDAR}      & 77.82  & 64.17  & 57.51 &66.5 &87.95 &78.39 &71.29 &79.21 & 220\\
         & ContFuse~\cite{CONTFUSE}         &ECCV 2018   &{LiDAR+RGB}    & 83.68  & 68.78  & 61.67 &71.38 &94.07 &85.35 &75.88 & 85.1 & 60\\
         & SECOND~\cite{yan2018second}      &Sensors 2018   &{LiDAR}      & 83.34  & 72.55  & 65.82 &73.9 &89.39 &83.77 &78.59 &83.92 & 50\\
         & PointPillars~\cite{lang2019pointpillars}&CVPR 2019   &{LiDAR}      & 82.58  & 74.31  & 68.99 &75.29 &90.07 &86.56 &82.81 &86.48 & \bf 23.6\\
         & TANet~\cite{liu2020tanet}        &AAAI 2020   &{LiDAR}      & 84.39  & 75.94  & 68.82 &76.38 &91.58  &86.54  &81.19  &86.44 & 34.75\\
         & Associate-3Ddet~\cite{du2020associate}&CVPR 2020   &{LiDAR}      & 85.99  & 77.40  & 70.53 &77.97 &91.40  &88.09  &82.96  &87.48 & 60\\
         & HotSpotNet~\cite{chen2019object}  &ECCV 2020   & {LiDAR}      &87.60 &78.31 &73.34 &79.75 &94.06 &88.09 &83.24 & 88.46 & 40*\\
         & Point-GNN~\cite{shi2020point}    &CVPR 2020   &{LiDAR}      & 88.33  & 79.47  & 72.29 &80.03 &93.11  &89.17  &83.90  &88.73 & 643\\
         & 3DSSD~\cite{yang20203dssd}       &CVPR 2020   &{LiDAR}      & 88.36  & 79.57  & 74.55 &80.83 &92.66  &89.02  &85.86  &89.18 & 38\\
         & SA-SSD~\cite{he2020structure}    &CVPR 2020    &{LiDAR}      & 88.75  & 79.79  & 74.16 &80.90 &95.03 &91.03 &85.96 &90.67 & 40.1\\
         & CIA-SSD~\cite{zheng2020cia}    &AAAI 2021    &{LiDAR}      & 89.59   & 80.28   & 72.87  &80.91 &93.74 &89.84 &82.39  &88.66 & 30.76\\
         \cline{2-12}
         & \bf SE-SSD (ours) &- &{LiDAR} & \bf91.49  &\bf82.54  &\bf77.15 &\bf83.73 &\bf95.68 &\bf91.84 &\bf86.72 &\bf91.41 & 30.56\\
      \hline
   \end{tabular}
   \vspace*{1mm}
   \caption{Comparison with the state-of-the-art methods on the KITTI \textit{test} set for car detection, with 3D and BEV average precisions of 40 sampling recall points evaluated on the KITTI server. Our SE-SSD attains the highest precisions for all difficulty levels with a very fast inference speed, outperforming all prior detectors. ``*'' means the runtime is cited from the submission on the KITTI website.
   }
   \vspace*{-3mm}
   \label{table1}
\end{table*}

\begin{table}[t]
\centering
\resizebox{\linewidth}{!}{
\begin{tabular}{@{\hspace{1mm}}c@{\hspace{1mm}}|@{\hspace{1mm}}c@{\hspace{1mm}}c@{\hspace{1mm}}c@{\hspace{1mm}}|@{\hspace{1mm}}c@{\hspace{1mm}}c@{\hspace{1mm}}c@{\hspace{1mm}}|@{\hspace{1mm}}c@{\hspace{1mm}}}
    \hline
      \multirow{2}{*}{Method}&
        \multicolumn{3}{@{\hspace{1mm}}c@{\hspace{1mm}}|@{\hspace{1mm}}}{3D$_{R40}$}
         &\multicolumn{3}{@{\hspace{1mm}}c@{\hspace{1mm}}|@{\hspace{1mm}}}{BEV$_{R40}$}
         &\multicolumn{1}{@{\hspace{1mm}}c@{\hspace{1mm}}}{3D$_{R11}$}\\
       \multicolumn{1}{@{\hspace{1mm}}c|@{\hspace{1mm}}}{} & Easy & Moderate & Hard & Easy & Moderate & Hard & Moderate\\ \hline \hline
        3DSSD~\cite{yang20203dssd}     & -        & -         & -       & -        & -           & -     &  79.45 \\
        SA-SSD~\cite{he2020structure}  & 92.23    & 84.30     & 81.36   & -        & -           & -     &  79.91 \\
        De-PV-RCNN~\cite{2020deformable}&-& 84.71     & -       & -        & -           & -     &  83.30 \\
        PV-RCNN~\cite{shi2020pv}       & 92.57    & 84.83     & 82.69   & 95.76    & 91.11       & 88.93 &  83.90 \\\hline
    \textbf{SE-SSE (ours)} &\textbf{93.19} &\textbf{86.12}&\textbf{83.31}&\textbf{96.59}&\textbf{92.28}&\textbf{89.72}&\textbf{85.71}\\ \hline
\end{tabular}
}
\vspace*{0mm}
\caption{Comparison with latest best two single- and two-stage detectors on KITTI \textit{val} split for car detection, in which ``R40'' and ``R11'' mean 40 and 11 sampling recall points for AP, respectively. }
\label{table3}
\vspace{-2.5mm}
\end{table}

\section{Experiments}
We evaluate our SE-SSD on the KITTI 3D and BEV object detection benchmark~\cite{geiger2013vision}.
This widely-used dataset contains 7,481 training samples and 7,518 test samples.
Following the common protocol, we further divide the training samples into a training set (3,712 samples) and a validation set (3,769 samples).
Our experiments are mainly conducted on the most commonly-used car category and evaluated by the average precision with an IoU threshold 0.7.
Also, the benchmark has three difficulty levels in the evaluation: easy, moderate, and hard, based on the object size, occlusion, and truncation levels, in which the moderate average precision is the official ranking metric for both 3D and BEV detection on the KITTI website.
We will {\em release our code on GitHub\/} upon the publication of this work.

Figure~\ref{detresults} shows 3D bounding boxes (2nd \& 5th rows) and BEV bounding boxes (3rd \& 6th rows) predicted by our SE-SSD model for six different inputs, demonstrating its high-quality prediction results.
Also, for a better visualization of the results, we project and overlay the 3D predictions onto the corresponding images (1st \& 4th rows).
Please refer to the supplemental material for more experimental results.

\subsection{Implementation Details}
\textbf{Data preprocessing}
We only use LiDAR point clouds as input and voxelize all points in ranges [0, 70.4], [-40, 40], and [-3, 1] meters into a grid of resolution [0.05, 0.05, 0.1] along $x$, $y$, and $z$, respectively.
We empirically set hyperparameters $p_1$=0.25, $p_2$=0.05, and $p_3$=0.1 (see Section~\ref{sec:3.2}).
Besides shape-aware data augmentation, we adopt three types of common data augmentation:
(i) mix-up~\cite{yan2018second}, which randomly samples ground-truth objects from other scenes and add them into the current scene;
(ii) local augmentation on points of individual ground-truth object,~\eg, random rotation and translation; and
(iii) global augmentation on the whole scene, including random rotation, translation, and flipping.
The former two are for preprocessing the inputs to both teacher and student SSDs.

\textbf{Training details}
We adopt the ADAM optimizer and cosine annealing learning rate~\cite{loshchilov2016sgdr} with a batch size of four for 60 epochs.
We follow~\cite{tarvainen2017mean} to ramp up $\mu_t$ (Eq.~\eqref{totalloss}) from 0 to 1 in the first 15 epoches using a sigmoid-shaped function $e^{-5(1-x)^2}$.
We set
$\tau_{c}$ and ${\tau_{I}}$ (Section~\ref{sec:3.3}) as 0.3 and 0.7, respectively,
$\omega_1$ and $\omega_2$ (Eq.~\eqref{totalloss}) as 2.0 and 0.2, respectively,
the EMA decay weight as 0.999, and
$\gamma$ (Eq.~\eqref{diou}) as 1.25.

\subsection{Comparison with State-of-the-Arts}
By submitting our prediction results to the KITTI server for evaluation, we obtain the 3D and BEV average precisions of our model on the KITTI test set and compare them with the state-of-the-art methods listed in Table~\ref{table1}.

As shown in the table, our model ranks the $1^{st}$ place among all state-of-the-art methods for both 3D and BEV detections in all three difficulty levels.
Also, the inference speed of our model ranks the $2^{nd}$ place among all methods, about 2.6 times faster than the latest best two-stage detector Deformable PV-RCNN~\cite{2020deformable}.
In 3D detection, our one-stage model attains a significant improvement of 1.1 points on moderate AP compared with PV-RCNN~\cite{2020deformable} and Deformable PV-RCNN~\cite{shi2020pv}.
For single-stage detectors, our model also outperforms all prior works by a large margin, outperforming the previous one-stage detector SA-SSD~\cite{he2020structure} by an average of 2.8 points for all three difficulty levels and with shorter inference time (reduced by $\sim$25$\%$).
Our large improvement in APs comes mainly from a better model optimization by exploiting both soft and hard targets, and the high efficiency of our model is mainly due to the nature of our proposed methods,~\ie, we refine features in SSD without incurring extra computation in the inference.

In BEV detection, our model also leads the best single- and two-stage detectors by around 0.8 points on average.
Besides, we calculate the mean average precision (mAP) of three difficulty levels for comparison.
Our higher mAPs indicate that SE-SSD attains a more balanced performance compared with others, so our method is more promising to address various cases more consistently in practice.
Further, we compare our SE-SSD with latest best two single- and two-stage methods on KITTI \textit{val} split.
As shown in Table~\ref{table3}, our 3D and BEV moderate APs with 11 or 40 recall points both outperform these prior methods.

\subsection{Ablation Study}
Next, we present ablation studies to analyze the effectiveness of our proposed modules in SE-SSD on KITTI \textit{val} split.
Table~\ref{table4} summarizes the ablation results on our consistency loss (``Cons loss''), ODIoU loss (``ODIoU''), and shape-aware data augmentation (``SA-DA'').
For ODIoU loss, we replace it with the Smooth-$L_1$ loss in this ablation study, since we cannot simply remove it like Cons loss and SA-DA.
All reported APs are with 40 recall points.

\begin{table}[t]
\centering
\resizebox{0.98\columnwidth}{!}{
\begin{tabular}{ccc|ccc}
    \hline
      \multicolumn{1}{c}{Cons loss} &\multicolumn{1}{c}{ODIoU} &\multicolumn{1}{c|}{SA-DA} &\multicolumn{1}{c}{ \multirow{1}{*}{Easy}} &\multicolumn{1}{c}{ \multirow{1}{*}{Moderate}} &\multicolumn{1}{c}{ \multirow{1}{*}{Hard}}\\
      \hline\hline
         -        &   -        &   -        & 92.58    & 83.22       & 80.15    \\
         -        &   -        & \checkmark & 93.02    & 83.70       & 80.68   \\
         -        & \checkmark &   -        & 93.07    & 83.85       & 80.78   \\
       \checkmark &   -        &   -        & 93.13    & 84.15       & 81.17   \\
       \checkmark & \checkmark &   -        & 93.17    & 85.81       & 83.01   \\
       \checkmark & \checkmark & \checkmark & \textbf{93.19}  & \textbf{86.12}  & \textbf{83.31}\\ \hline
\end{tabular}
}
\vspace*{0mm}
 \caption{Ablation study on our designed modules.
 We report the 3D average precisions of 40 sampling recall points on KITTI \textit{val.}~split for car detection.
 Cons Loss and SA-DA denote our consistency loss and shape-aware data augmentation, respectively.
 }
\label{table4}
\vspace{-1.5mm}
\end{table}

\textbf{Effect of consistency loss}
As first and fourth rows in Table~\ref{table4} show, our consistency loss boosts the moderate AP by about 0.9 point.
This large improvement shows that using the more informative soft targets can contribute to a better model optimization.
For the slight increase in easy AP, we think that the predictions of the baseline on the easy subset are already very precise and thus are very close to the hard targets already.
Importantly, by combining hard labels with the ODIoU loss in the optimization, our SE-SSD further boosts the moderate and hard APs by about 2.6 points, as shown in the fifth row in Table~\ref{table4}.
This demonstrates the effectiveness of jointly optimizing the model by leveraging both hard and soft targets with our designed constraints.

Further, we analyze the effect of the consistency loss for bounding boxes (``reg'') and confidence  (``cls'') separately to show the effectiveness of the loss on both terms.
As Table~\ref{table5} shows, the gain in AP from the confidence term is larger, we argue that the confidence optimization may be more effective to alleviate the misalignment between the localization accuracy and classification confidence.
Also, we evaluate the box and confidence constraints~\cite{zhao2020sess} designed on the box-centers distance matching strategy and obtain a much lower AP (``dist''), we think that the underlying reason is related to their design, which is to address axis-aligned boxes and so is not suitable for our task.

\textbf{Effect of ODIoU loss}
As first and third rows in Table~\ref{table4} show, our ODIoU loss improves the moderate AP by about 0.6 points compared with the Smooth-$L_1$ loss.
This gain in AP is larger than the one with the SA-DA module, thus showing the effectiveness of the constraints on both the box-centers distance and orientation difference in the ODIoU loss.
Also, the gain in AP on the hard subset is larger than others, which is consistent with our expectation that even sparse points on the object surface could provide sufficient information to regress the box centers and orientations.

\textbf{Effect of shape-aware data augmentation}
In Table~\ref{table4}, the first two rows indicate that our shape-aware data augmentation (SA-DA) scheme brings an average improvement of about 0.5 points on the baseline model.
Based on the pre-trained SSD,  SA-DA further improves the moderate and hard APs of SE-SSD by about 0.3 points, as indicated in the last two rows in Table~\ref{table4}.
These gains in AP show the effectiveness of our SA-DA on boosting the performance by enhancing the object diversity and model generalizability.

\textbf{IoU-Based Matching Strategy}
Also, we compare different ways of filtering soft targets,~\ie, removing soft targets that
(i) overlap with each other using NMS (``nms filter''),
(ii) do not overlap with any ground truth (``gt filter''), and
(iii) do not overlap with student boxes for less than an IoU threshold (``stu filter'').
We can see from Table~\ref{table6} that our proposed ``stu filter'' attains the largest gain in AP, as it keeps the most related and informative soft targets for the student predictions, compared with other strategies.

\begin{table}[t]
\small
\centering
\resizebox{\linewidth}{!}{
\begin{tabular}{@{\hspace{1mm}}c@{\hspace{1.5mm}}|@{\hspace{1.5mm}}c@{\hspace{2mm}}c@{\hspace{2mm}}c@{\hspace{2mm}}c@{\hspace{2mm}}c@{\hspace{1mm}}}
    \hline
       Type         &baseline   & dist   & reg only      & cls only      & cls + reg                   \\ \hline
       Moderate AP  &83.22      & 80.38      & 83.65         & 83.83         & \textbf{84.15}                \\  \hline
\end{tabular}
}
\vspace*{0.25mm}
 \caption{Ablation study on our consistency loss, in which ``cls'' and ``reg'' mean our consistency loss on confidence and bounding boxes, respectively, and ``dist'' means the box and confidence constraints based on a box-centers distance matching strategy.}
\label{table5}
\vspace{-1.5mm}
\end{table}

\begin{table}[t]
\small
\centering
\resizebox{\linewidth}{!}{
\begin{tabular}{@{\hspace{1mm}}c|c@{\hspace{3mm}}c@{\hspace{3mm}}c@{\hspace{3mm}}c@{\hspace{1mm}}}
    \hline
       Type          &baseline  & nms filter      & gt filter         & stu filter             \\ \hline
       Moderate AP   &83.22     &  83.49          & 80.73             & \textbf{84.15}           \\  \hline
\end{tabular}
}
\vspace*{0.25mm}
 \caption{Ablation study on our IoU-based matching strategy, in which ``nms'', ``gt'', and ``stu'' mean that we filter soft targets with NMS, ground truths, and student predictions, respectively.}
\label{table6}
\vspace{-1.5mm}
\end{table}

\subsection{Runtime Analysis}
The overall inference time of SE-SSD is only 30.56ms, including
2.84ms for data preprocessing,
24.33ms for network forwarding, and
3.39ms for post-processing and producing the final predictions.
All evaluations were done on an Intel Xeon Silver CPU and a single TITAN Xp GPU.
Our method attains a faster inference speed compared with SA-SSD~\cite{he2020structure}, as our BEVConvNet structure is simpler and we further optimized our voxelization code.

\ifx\allfiles\undefined
\end{document}
\fi 
\ifx\allfiles\undefined

\documentclass[letterpaper]{article}
\begin{document}
\else
\chapter{Conclusion}
\fi

\section{Conclusion}

This paper presents a novel self-ensembling single-stage object detector for outdoor 3D point clouds.
The main contributions include the SE-SSD framework optimized by our formulated consistency constraint with soft targets, the ODIoU loss to supervise the network with hard targets, and the shape-aware data augmentation scheme to enlarge the diversity of training samples.
The series of experiments demonstrate the effectiveness of SE-SSD and each proposed module, and show the advantage of high efficiency.
Overall, our SE-SSD outperforms all state-of-the-art methods for both 3D and BEV car detection in the KITTI benchmark and attains an ultra-high inference speed.

\vspace*{1mm}
\noindent
{\bf Acknowledgments.} \
This work is supported by the Hong Kong Centre for Logistics Robotics.

\ifx\allfiles\undefined
\end{document}
\fi 
{\small
\bibliographystyle{ieee_fullname}
\bibliography{egbib}
}
\clearpage
\clearpage
\ifx\allfiles\undefined
\documentclass[letterpaper]{article}
\begin{document}
\else
\chapter{paper_supp}
\fi

This supplementary material contains the following three sections.
Section~\ref{sec:C} shows screenshots of the KITTI 3D and BEV leaderboard taken on the date of CVPR submission, showing the rank and time performance of our SE-SSD.
Section~\ref{sec:A} presents further ablation studies to analyze our ODIoU loss and shape-aware data augmentation on KITTI car dataset.
Section~\ref{sec:B} shows the 3D and BEV detection results of our baseline SSD and SE-SSD on the KITTI cyclist and pedestrian benchmarks.
All our results on the KITTI \textit{val} split are averaged from multiple runs and evaluated with the average precision of 40 sampling recall points.

\appendix


\section{KITTI Car Detection Leaderboards}\label{sec:C}
\begin{figure}[h]
\centering
\vspace*{-2mm}
\includegraphics[width=8cm]{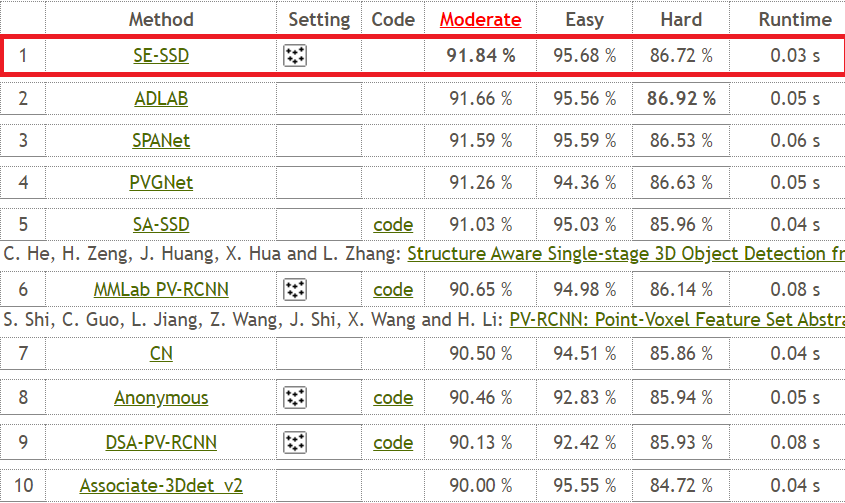}
\caption{KITTI BEV (Bird's Eye View) car detection leaderboard, in which our SE-SSD ranks the $1st$ place.}
\label{fig:BEVleaderboard}
\vspace*{-3mm}
\end{figure}

\begin{figure}[h]
\centering
\includegraphics[width=8cm]{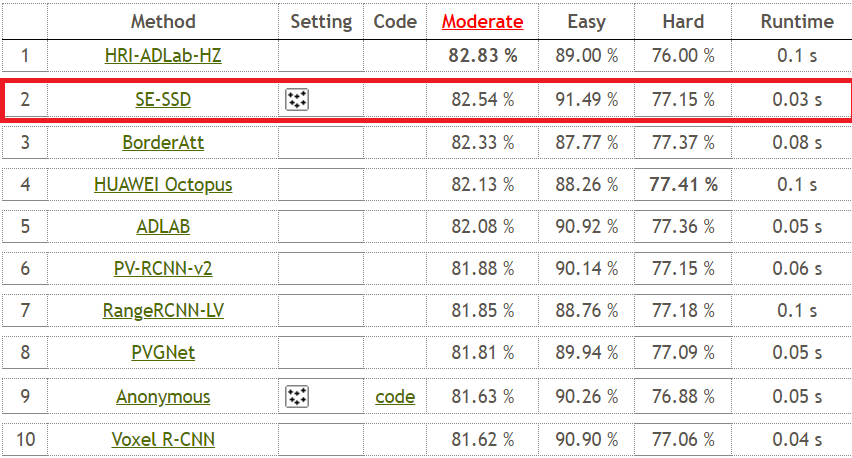}
\caption{KITTI 3D car detection leaderboard, in which our SE-SSD ranks the $2nd$ place (HRI-ADLab-HZ is unpublished).}
\label{fig:3dleaderboard}
\vspace*{-2mm}
\end{figure}

As shown in Figures~\ref{fig:BEVleaderboard} and~\ref{fig:3dleaderboard}, our SE-SSD ranks $1st$ and $2nd$ on the KITTI BEV and 3D leaderboards of car detection~\footnote{On the date of CVPR deadline,~\ie, Nov 16, 2020}, respectively, comparing with not only prior published works but also unpublished works submitted to the leaderboard.
Also, our SE-SSD runs the fastest among the top submissions and achieved a balanced performance for the three difficulty levels, especially in 3D detection.

\section{More Ablation Studies}\label{sec:A}

\begin{table}[!t]
\small
\centering
\resizebox{\linewidth}{!}{
\begin{tabular}{@{\hspace{1mm}}c@{\hspace{1.5mm}}|@{\hspace{1.5mm}}c@{\hspace{2mm}}c@{\hspace{2mm}}c@{\hspace{2mm}}c@{\hspace{2mm}}c@{\hspace{1mm}}}
    \hline
       Type         &baseline   & dropout   & swap      & sparsify     & Full SA-DA                   \\ \hline
       Moderate AP  &83.22      & 83.46     & 83.48     & 83.43        & \textbf{83.70}          \\  \hline
\end{tabular}
}
\vspace*{0.25mm}
 \caption{Ablation study on the operators (random dropout, swap, and sparsifying) in our shape-aware data augmentation (SA-DA).}
\label{table0_sup}
\end{table}

\noindent
\textbf{Shape-aware data augmentation}
We analyze the effect of random dropout, swap, and sparsifying in our shape-aware data augmentation on KITTI \textit{val} split for car detection, respectively.
As Table~\ref{table0_sup} shows, all these operators (random dropout, swap, and sparsifying) boost the 3D moderate AP effectively, thus demonstrating the effectiveness of our proposed augmentation operators to enrich the object diversity.

\begin{table}[!t]
\small
\centering
\resizebox{\linewidth}{!}{
\begin{tabular}{@{\hspace{1mm}}c@{\hspace{1.5mm}}|@{\hspace{1.5mm}}c@{\hspace{2mm}}c@{\hspace{2mm}}
c@{\hspace{2mm}}c@{\hspace{2mm}}c@{\hspace{2mm}}c@{\hspace{2mm}}c@{\hspace{1mm}}}
    \hline
       $\gamma$        & 0.25    & 0.5      & 0.75     & 1.0     & 1.25           & 1.5    & 1.75      \\ \hline
       Moderate AP     & 83.47   & 83.65    & 83.73    & 83.78   & \textbf{83.85} & 83.58  & 83.52         \\  \hline
\end{tabular}
}
\vspace*{0.25mm}
 \caption{Ablation study on our ODIoU loss, in which we compare the 3D moderate AP of different settings of $\gamma$.}
\label{table1_sup}
\end{table}

\textbf{ODIoU Loss}
Next, we try different values of $\gamma$ in the ODIoU loss on KITTI \textit{val} split for car detection.
As Table~\ref{table1_sup} shows, the orientation constraint is an important factor to further boost the precision, so we finally set $\gamma$ as 1.25.

\section{Experiments on KITTI Cyclist$\&$Pedestrian}\label{sec:B}
To validate the effectiveness of our SE-SSD framework, we further conduct experiments on the Cyclist and Pedestrian datasets in KITTI benchmark.
In Tables~\ref{table2_sup} and~\ref{table3_sup}, we compare the 3D and BEV average precisions between the baseline SSD and our SE-SSD on KITTI \textit{val} split.

\begin{table}[!t]
        \small
        \centering
        \begin{tabular}{@{\hspace{1mm}}cc|ccc@{\hspace{1mm}}}
            \hline
            \multicolumn{2}{c|}{Cyclist}  &\multicolumn{1}{c}{ \multirow{1}{*}{ Easy}} &\multicolumn{1}{c}{ \multirow{1}{*}{ Moderate}} &\multicolumn{1}{c}{ \multirow{1}{*}{ Hard}} \\
            \hline\hline
            \multirow{2}{*}{3D}     & SSD            &  75.73           &  55.86          & 51.97  \\
                                    & our SE-SSD         &  \textbf{80.07}  & \textbf{70.43}  &\textbf{66.45}\\
            \hline
            \multirow{2}{*}{BEV}    & SSD            &  83.71           &  59.02          &  55.05 \\
                                    & our SE-SSD          &\textbf{91.83}    & \textbf{72.62}  &\textbf{68.24}\\
            \hline
        \end{tabular}
\vspace*{2mm}
    \caption{Comparison of 3D and BEV APs between our baseline SSD and SE-SSD on KITTI \textit{val} split for ``cyclist'' detection.}
    \label{table2_sup}
\end{table}

\begin{table}[!t]
\scriptsize
\small
        \centering
        \begin{tabular}{@{\hspace{1mm}}cc|ccc@{\hspace{1mm}}}
            \hline
            \multicolumn{2}{c|}{Pedestrian}  &\multicolumn{1}{c}{ \multirow{1}{*}{ Easy}} &\multicolumn{1}{c}{ \multirow{1}{*}{ Moderate}} &\multicolumn{1}{c}{ \multirow{1}{*}{ Hard}} \\
            \hline\hline
            \multirow{2}{*}{3D}     & SSD             &  59.64           &  52.63          & 46.59  \\
                                    & our SE-SSD          &  \textbf{63.27}  & \textbf{57.32}  &\textbf{50.82}\\
            \hline
            \multirow{2}{*}{BEV}    & SSD             &  63.53           &  57.29          &  51.36 \\
                                    & our SE-SSD          &\textbf{67.47}    & \textbf{61.88}  &\textbf{55.94}\\
            \hline
        \end{tabular}
    \vspace*{2mm}
    \caption{Comparison of 3D and BEV APs between our baseline SSD and SE-SSD on KITTI \textit{val} split for ``pedestrian'' detection.}
    \label{table3_sup}
\end{table}

\textbf{Cyclist \& Pedestrian Results}
As Table~\ref{table2_sup} shows, our SE-SSD outperforms the baseline SSD by a large margin for both 3D and BEV cyclist detection, especially on the 3D moderate and hard subsets with an improvement of about 15 points.
As Table~\ref{table3_sup} shows, our SE-SSD also outperforms the baseline SSD on both the 3D and BEV pedestrian detection by a large margin.
These large improvements clearly show the effectiveness of our proposed SE-SSD framework.

\ifx\allfiles\undefined
\end{document}
\fi

\end{document}